\documentclass[sigconf]{acmart}

\AtBeginDocument{%
  \providecommand\BibTeX{{%
    \normalfont B\kern-0.5em{\scshape i\kern-0.25em b}\kern-0.8em\TeX}}}

\setcopyright{acmcopyright}

\begin{document}

\copyrightyear{2019}
\acmYear{2019}
\acmConference[MM '19]{Proceedings of the 27th ACM International Conference on Multimedia}{October 21--25, 2019}{Nice, France}
\acmBooktitle{Proceedings of the 27th ACM International Conference on Multimedia (MM '19), October 21--25, 2019, Nice, France}
\acmPrice{15.00}
\acmDOI{10.1145/3343031.3351083}
\acmISBN{978-1-4503-6889-6/19/10}

\title{M2E-Try On Net: Fashion from Model to Everyone}

\author{Zhonghua Wu}
\email{zhonghua001@e.ntu.edu.sg}
\affiliation{
 \institution{Nanyang Technological University}}
 
 \author{Guosheng Lin}
\email{gslin@ntu.edu.sg}
\affiliation{
 \institution{Nanyang Technological University}}

 \author{Qingyi Tao}
\email{qtao002@e.ntu.edu.sg}
\affiliation{
 \institution{Nanyang Technological University \and NVIDIA AI Technology Center}}
 
  \author{Jianfei Cai}
  \affiliation{
 \institution{Nanyang Technological University}}
\email{Jianfei.cai@monash.edu}
\affiliation{
 \institution{Monash University}}

\begin{abstract}
Most existing virtual try-on applications require clean clothes images. Instead, we present a novel virtual Try-On network, M2E-Try On Net, which transfers the clothes from a model image to a person image without the need of any clean product images. To obtain a realistic image of person wearing the desired model clothes, we aim to solve the following challenges: 1) non-rigid nature of clothes - we need to align poses between the model and the user; 2) richness in textures of fashion items - preserving the fine details and characteristics of the clothes is critical for photo-realistic transfer; 3) variation of identity appearances - it is required to fit the desired model clothes to the person identity seamlessly. To tackle these challenges, we introduce three key components, including the pose alignment network (PAN), the texture refinement network (TRN) and the fitting network (FTN). Since it is unlikely to gather image pairs of input person image and desired output image (i.e. person wearing the desired clothes), our framework is trained in a self-supervised manner to gradually transfer the poses and textures of the model's clothes to the desired appearance. In the experiments, we verify on the Deep Fashion dataset and MVC dataset that our method can generate photo-realistic images for the person to try-on the model clothes. Furthermore, we explore the model capability for different fashion items, including both upper and lower garments.

\end{abstract}



\begin{teaserfigure}
  \includegraphics[width=\textwidth]{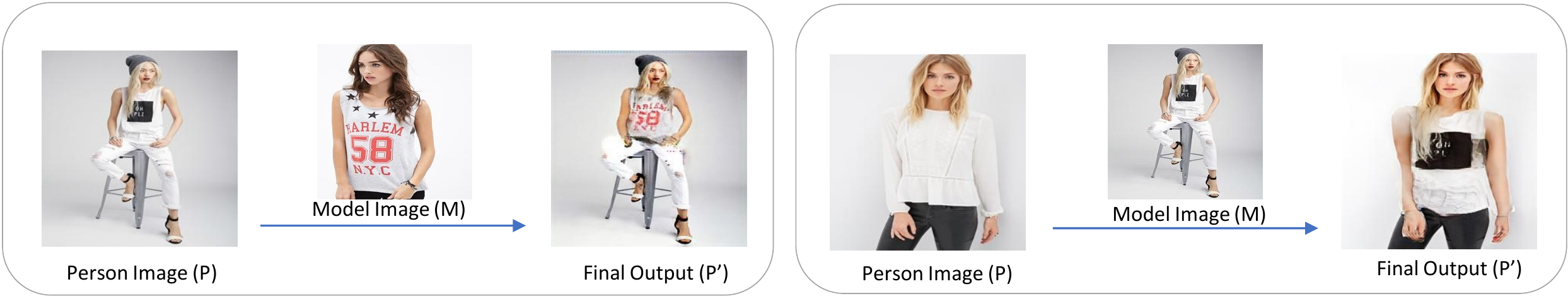}
  \caption{M2E-TON can try-on the clothes from model $M$ to person $P$ to generate a new person image $P'$ with desired model clothes without clean product images. }
  \label{first_diagram}
\end{teaserfigure}

\maketitle

\begin{figure*}[t]
    \includegraphics[width=\textwidth]{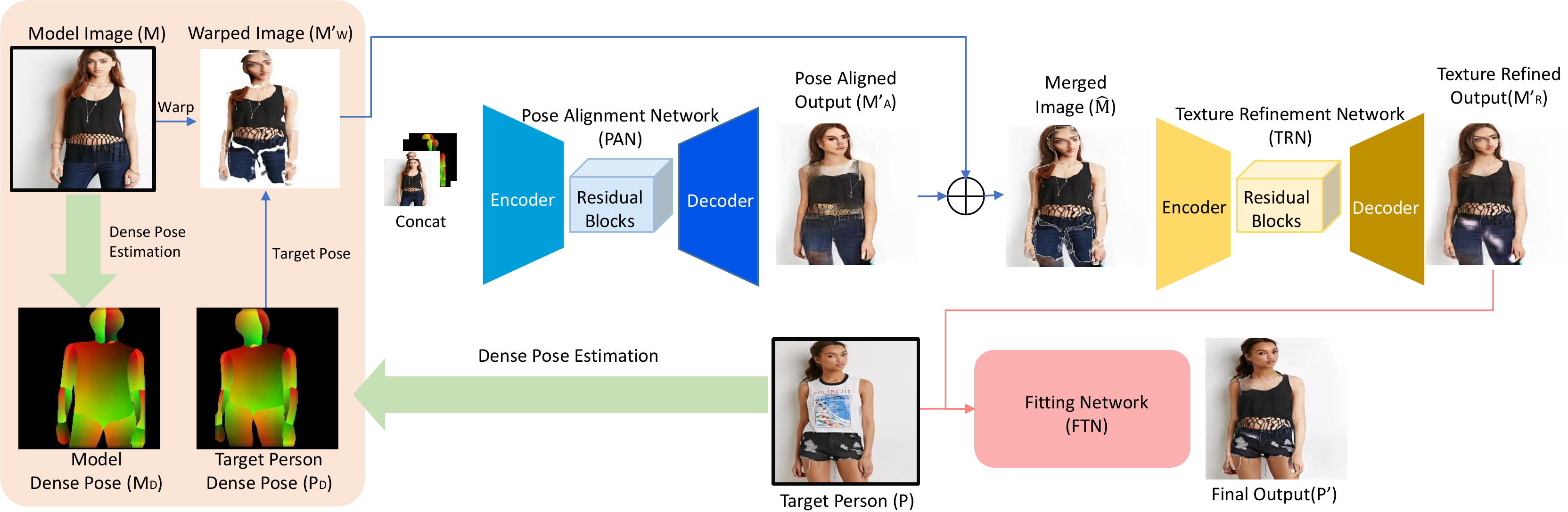}
    \caption{Overall pipeline of our M2E-TON. Our network contains three stages. The inputs of our proposed M2E-TON are only Model Image ($M$) and Target Person image ($P$). The left one (blue) is the first stage called Pose Alignment Network (PAN), which could align model image $M$ into $M'_A$ whose pose is same as the target person pose based on four inputs. The right one (yellow) is a refined stage called Texture Refinement Network (TRN) which could refine the merged image $\hat{M}$. The bottom one (pink) is the Fitting Network (FTN), which would transfer the desired clothes into the person image to generate final output $P'$.}
    \label{overall_diagram}
\end{figure*}

\section{Introduction}
With the emergence of Internet marketing and social media, celebrities and fashion bloggers like to share their outfits online. While browsing those adorable outfits, people may desire to try out these fashion garments by themselves. In this scenario, a virtual try-on system allows the customers to change the clothes immediately, without any pain of inconvenient travel, by automatically transferring the desired outfits to their own photos. Such garment transferring application can be adopted by the e-commerce platforms as the virtual ``fitting room'', or photo editing tools for designing or entertainment purposes.

Earlier virtual try-on systems often relies on 3D information of body shape, and even the garment \cite {niswar2011virtual, sekine2014virtual, yang2016detailed}, which is unrealistic for normal users and for every-day use. Thus, pure image-based methods proposed recently have attracted extensive attention \cite{jetchev2017conditional,kubo2018generative,han2017viton,wang2018toward}. In \cite{han2017viton, wang2018toward}, they deform clothes from clean product images by applying the geometric transformation and then fit the deformed clothes to the original person images. However, these methods require clean product image of the desired clothes, which are not always available and accessible to the users, especially for those fashion items shared on social media.

In this work, we introduce a virtual try-on approach that can transfer the clothes from any forms of photos (e.g. the model shot images or photos from social media posts) to the target identities (i.e. the users), without relying on clean product images. As illustrated in Figure \ref{first_diagram}, in our approach, a person $P$ can try-on the clothes worn by a model $M$ and the output image is generated in which the target identity is wearing the desired clothes from the model, denoted as $P'$ while the personal identity and the clothes characteristics are well preserved.

A common challenge of the virtual try-on solution is the absence of paired input image (original person image) and output image (ground truth image of the person with the desired clothes). In this circumstance, existing works \cite{jetchev2017conditional,kubo2018generative} exploit the popular Cycle-GAN \cite{zhu2017unpaired} framework with cycle consistency between cycle-generated image and the original image to achieve the clothes swap task. However, due to the weak constraints within the cycle, these methods suffer from degraded details in textures, logo patterns and styles, making them prone to being unrealistic. In contrast, to break the cycled design, our method incorporates a joint learning scheme with unsupervised training on unpaired images and self-supervised training on paired images of the same identity (wearing the same clothes but having different poses).

Importantly, unlike the above mentioned methods \cite{jetchev2017conditional,kubo2018generative,han2017viton,wang2018toward}, since our approach focuses on the virtual try-on directly from the model image rather than the clean product image, it is more challenging to fit the garment from the model image to the person image seamlessly, so as to well conserve the target person appearance.

Specifically, our ``Model to Everyone - Try On Net'' (M2E-TON) includes three major components: the pose alignment network (PAN) to align the model and clothes pose to the target pose, the texture refinement network (TRN) to enrich the textures and logo patterns to the desired clothes, and the fitting network (FTN) to merge the transferred garments to the target person images. With the unpaired-paired joint training scheme, the proposed M2E-TON is able to transfer desired model clothes to the target person image automatically.

We experiment on Deep Fashion \cite{liu2016deepfashion} Women Tops dataset, MVC \cite{liu2016mvc} Women Tops dataset and MVC \cite{liu2016mvc} Women Pants dataset and show that we can achieve virtual try-on task on various poses of persons and clothes. Apart from tops and pants, the proposed method is scalable to other body parts, such as hair, shoes, gloves and etc.

The main contributions of our work are summarized as follows:
\begin{itemize}
\item We propose a virtual Try-On Network named M2E-TON that can transfer desired model clothes to arbitrary person images automatically. Through our carefully designed sub-networks, our model can generate photo-realistic results with aligned poses and well-preserved textures. Specifically, our proposed method does not need desired product images which are typically unavailable or inaccessible.  

\item We propose three sub-networks for our virtual try-on model: 1) We introduce a pose alignment network (PAN) to align model image to the target person pose; 2) To enhance the characteristics of the clothes, we propose a Texture Refinement Network (TRN) to add the transformed clothes textures to the pose aligned image; 3) Lastly, we utilize a Fitting Network (FTN) to fit the desired clothes to the original target person image.

\item Due to the lack of paired training data (the target person $P$ with desired model clothes, denoted as $P'$ in Figure \ref{first_diagram}), we propose an innovative hybrid learning framework of unsupervised learning and self-supervised learning with pose conditional GAN to accomplish this task.

\end{itemize}

\begin{figure*}[t]
 \centering
    \includegraphics[width=\textwidth]{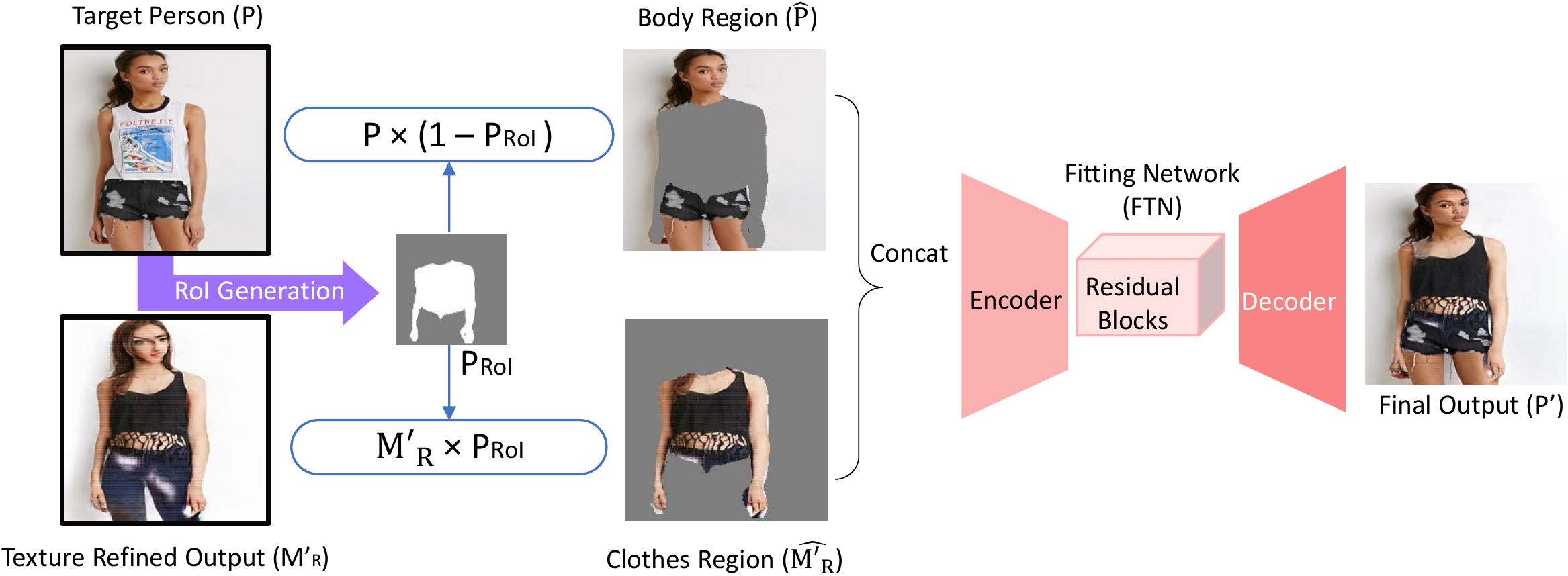}
    \caption{Illustration of the Fitting Network (FTN). Firstly, a Region of Interest (RoI) is generated by the RoI Generation Network. Based on the obtained RoI, we use FTN to merge the transferred garment image $M'_R$ and original person image $P$.}
    \label{stage_three}
\end{figure*}

\section{Related Works}

Our method is related to the research topics including human parsing and analysis, person image generation and virtual try-on and fashion datasets.

\subsection{Human Parsing and Human Pose Estimation}
Human parsing has been studied for fine-grained segmentation of human body parts \cite{wu2018keypoint, Lin:2017:RefineNet}. Chen et al. \cite{chen_cvpr14} extended object segmentation to object part-level segmentation and released the PASCAL PART dataset including pixel-level part annotations of the human body. Gong et al. \cite{gong2017look} collected a new person dataset (LIP) for human parsing and fashion clothes segmentation. Lin et al. \cite{Lin:2017:RefineNet} proposed a multi-path refinement network to achieve high resolution and accurate part segmentation. Our work exploits \cite{gong2017look} to extract the region of interest that covers the clothes part in person images.

Apart from human parsing for part segmentation, the works in \cite{cao2017realtime,guler2018densepose} studied human pose estimation for pose analysis. Cao et al. \cite{cao2017realtime} proposed a Part Affinity Fields for human pose estimation based on key points. Later, to achieve more accurate pose estimation, DensePose \cite{guler2018densepose} proposed dense human pose estimation method by mapping each pixel to a dense pose point. In our work, we utilize the estimated dense poses for clothes region warping and pose alignment. 

\subsection{Person Image Generation and Virtual Try-On}
Generative adversarial network (GAN) \cite{goodfellow2014generative} has been used for image-based generation.
Recently, GAN has been used for person image generation \cite{ma2017pose} to generate the human image from pose representation. Zhu et al. \cite{zhu2017your} proposed a generative network to generate fashion images from textual inputs. For fashion image generation, a more intuitive way is to generate images from a person image and the desired clothes image, i.e. a virtual try-on system. This virtual try-on task has been studied in the past a few years \cite{zhu2017unpaired,jetchev2017conditional,han2017viton,wang2018toward}. Inspired by cycle image translation \cite{zhu2017unpaired}, Jetchev et al. \cite{jetchev2017conditional} proposed to translate person clothes using product clothes as the condition. To preserve a detailed texture information, Han et al. \cite{han2017viton} proposed to generate a synthesized image from clothes image and clothing-agnostic person representation. Wang et al. \cite{wang2018toward} proposed to refine the clothes detail information by adding a warped product image by applying Geometric Matching Module (GMM). However, these methods rely on product images as the input. Instead, our method focuses on transferring the dressed clothes on an arbitrary model image without  the need for clean product images. 

Similarly, some recent works \cite{neverova2018dense,raj2018swapnet,dong2018soft, liu2019swapgan, zanfir2018human, jiang2017fashion} have been proposed to transfer the person images to different variations with the corresponding human poses. Guler et al. \cite{guler2018densepose} proposed a pose transfer network based on dense pose condition. The methods in \cite{dong2018soft} and \cite{raj2018swapnet} use human part segmentation to guide the human pose translation. These methods are merely based on the generative models which often fail to recover the textures of garments. In contrast, our method is able to preserve the characteristics of clothes. At the same time, our method can transfer the garment from the model person to the target person while preserving the identity and pose.

\subsection{Fashion Datasets}

Our work uses several fashion datasets \cite{liu2016deepfashion,liu2016mvc} for training and evaluating our try-on network.
Deep Fashion dataset \cite{liu2016deepfashion} is a fashion dataset for clothes attribute prediction and landmark detection. MVC dataset \cite{liu2016mvc} is for view-invariant clothing retrieval and attribute prediction. These datasets provide us a large number of dressed person images with multiple views, poses and identities. 

\begin{figure*}[t]
    \includegraphics[width=\textwidth]{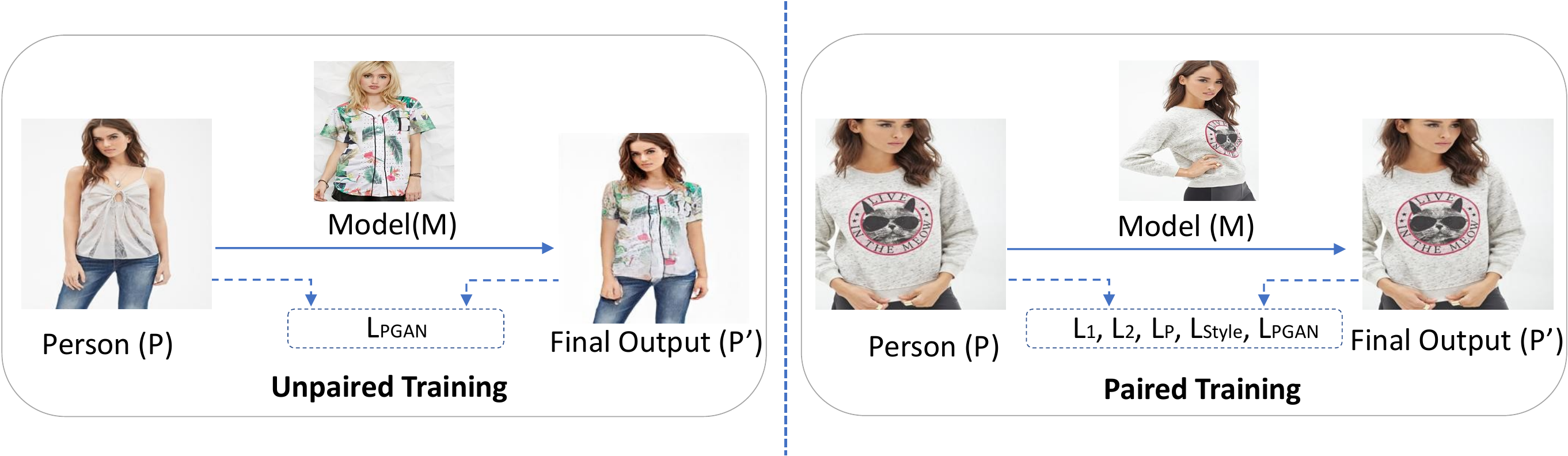}
    \caption{An illustration of our unpaired-paired Joint Training strategy. For the unpaired training, we only use GAN loss. Kindly note that there is no available ground truth output image in our unpaired training. For the paired training, we add a set of pixel-level similarity losses.}
    \label{pair_unpair}
\end{figure*}

\begin{figure*}[t]
 \centering
    \includegraphics[width=0.8\textwidth]{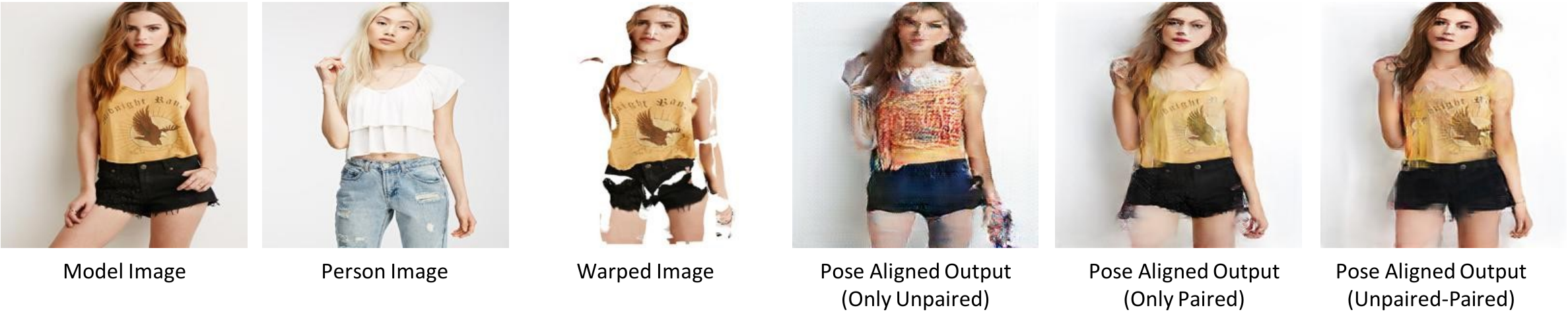}
    \caption{Effect of the unpaired-paired joint training for the Pose Alignment Network (PAN). The unpaired-paired joint training strategy is able to preserve both geometry and texture information compared with only unpaired or paired training scheme. }
    \label{show_unpair_pair}
\end{figure*}

\section{Approach}
We propose a human clothes try-on system which can virtually transfer clothes from model $M$ to person $P$ to generate a new person image $P^{'}$ in which he or she wears the clothes from the model $M$. 

As shown in Figure \ref{overall_diagram}, we use a three-stage approach to transfer the clothes from the model to the person. In particular, we propose a dense pose \cite{guler2018densepose} based human Pose Aligned Network (PAN) to transfer the model image to the target person pose. Then we propose a Texture Refinement Network (TRN) to merge the output images from PAN with the warped image. Finally, we propose a Fitting Network (FTN) to transfer the desired model clothes to the target person.

\subsection{Pose Alignment Network (PAN)}

As shown in Figure \ref{overall_diagram}, in the first stage of our pipeline, we develope a Pose Alignment Network (PAN) to align the pose of model person $M$ with target person $P$. The inputs of PAN consist of model image $M$, model dense pose $M_D$ generated from $M$, person dense pose $P_D$ generated from $P$ and model warped image $M'_W$ that is generated from $M$, $M_D$ and $P_D$. We consider PAN as a conditional generative module: 
it generates $M'_A$ from $M$ with conditions on $M_D$, $P_D$ and $M'_W$, where $M_D$ and $P_D$ are used to transfer the poses from model pose to target pose, and $M'_W$ provides strong condition on the clothes textures. 

Firstly, we use \cite{guler2018densepose} to generate dense poses of model ($M_D$) and target person images($P_D$). Each dense pose prediction has a partition of 24 parts. For each part, it has UV parametrization of the body surface. We use the UV coordinates to warp the model image $M$ to $M'_W$ through the dense pose prediction $M_D$ and $P_D$ by barycentric coordinate interpolation \cite{schindler2012barycentric}.

Secondly, given the inputs of $M$, $M_D$, $P_D$ and $M'_W$, the proposed Pose Alignment Network (PAN) learns to generate a pose aligned image $M'_A$ by transferring the model image $M$ to the target pose. In particular, we concatenate $M$, $M_D$, $P_D$ and $M'_W$ as the network input of 256 $\times$ 256 $\times$ 12. We utilize a multi-task encoder which contains three convolution layers to encode the 256 $\times$ 256 $\times$ 12 input to 64 $\times$ 64 $\times$ 256 feature representation, followed by a set of residual blocks with 3 $\times$ 3 $\times$ 256 $\times$ 256 kernels. Then, we use a decoder with two deconvolution layers and one convolution layer followed by a tanh activate function to reconstruct $M'_A$.

To train PAN, ideally we need to have a training triplet with paired images: model image $M$, person image $P$, and pose aligned model image $M'_A$ whose pose is the same as the person image $P$. However, it is difficult to obtain such training triplets with paired images. To ease this issue, we propose a self-supervised training method that uses images of the same person in two different poses to supervise Pose Alignment Network (PAN). We will further elaborate this in Section \ref{section pair}.

\subsection{Texture Refinement Network (TRN)}
Our Texture Refinement Network (TRN) is to use the warped image $M'_W$ to enhance the texture of the output image $M'_A$ from PAN, as shown in Figure \ref{overall_diagram}.

In order to ensure the details of the generated image, we follow a common practice to combine the information from network generated images and texture preserved images produced by geometric transformation. Particularly, we use the texture details from $M'_W$ to replace the corresponding pixel values in $M'_A$. In detail, we define the region of texture as $R$. $R$ is a binary mask with the same size as $M'_W$. If a pixel of $M'_W$ is not of the background color, its mask value equals to 1; otherwise 0.

Our merged image can be obtained as below:

\begin{equation}
\hat{M} = M'_W \odot R + M'_A \odot (1 - R), 
\label{energy}
\end{equation}
where $\odot$ indicates element-wise multiplication for each pixel. However, merged images are likely to have a sharp color change around the edge of $R$ which makes them unrealistic. Therefore, we propose to use a Texture Refinement Network (TRN) to deliver a texture refined image $M'_R$. This network helps smooth the merged images while still preserving the textual details on the garments. 


\subsection{Fitting Network (FTN)}
Given the refined apparel image, the last step of our approach is to fit the clothes to the target person image seamlessly. As shown in Figure \ref{stage_three}, we propose a Fitting Network to merge the transferred garment image $M'_R$ and original person image $P$. 

Firstly, to obtain the Region of Interest (RoI) for our fitting process, we use the LIP\_SSL pretrained network \cite{gong2017look} to generate the clothes mask and use DensePose \cite{guler2018densepose} estimation model to produce the upper body region mask. We then merge these two regions to a union mask since we are interested in transferring not only the garment regions but also body parts like the arms. To improve the smoothness of the RoIs, we train an RoI generation network using the union masks as pseudo ground truth and we use L1 loss as the cost function here.

After obtained the RoI masks (denoted as $P_{RoI}$), we apply the mask on both the texture refined image $M'_R$ and the person image $P$ as below:

\begin{equation}
\begin{cases}
\hat{M'_R} = M'_R \odot P_{RoI} ; \cr
\hat{P} = P \odot (1 - P_{RoI}) ,
\end{cases}
\end{equation}
where $\odot$ indicates element-wise multiplication on each pixel. The resulting $\hat{M'_R}$ and $\hat{P}$ are then concatenated as the network input of size 256 $\times$ 256 $\times$ 6. Similarly, our Fitting Network is an encoder-decoder network, including three convolution layers as the encoder, six residual blocks for feature learning, followed by two deconvolution layers and one convolution layer as the decoder. 

Again, similar to the learning of PAN, we can hardly find paired training data for this network, so we adopt the joint training scheme, which will be introduced in section \ref{section pair} to alleviate the need of supervision on the target outputs.

\subsection{Unpaired-Paired Joint Training}
\label{section pair}
As aforementioned, ideally we need to have paired training samples: model image $M$, person image $P$, and the ground truth image $P'$ where the person is wearing the model's clothes with exactly the same pose as in the original person image. In practice, these image pairs are unlikely to be obtained. Thus, we propose two learning methods to collaboratively train the network: 1) training with unpaired images and 2) training with paired images of the same identity (wearing the same clothes but having different poses). 

In particular, we first simply train our framework using random pairs of training images, i.e. unpaired images, using GAN loss, as shown in Figure~\ref{pair_unpair}. The discriminator learns to distinguish between the real person images and generated person images in order to guide generators to generate more realistic images. Nevertheless, since there is no strong supervision for the output images, there is lack of structural correspondence between the input and the desired output. On the other hand, we observed that in most fashion datasets, there are paired images of the same identity wearing the same clothes but with different poses. As shown in Figure~\ref{pair_unpair}, we propose a self-supervised paired training approach by training these ``paired'' images with additional pixel-level similarity losses (e.g. $\ell$1, $\ell$2).

In Figure ~\ref{show_unpair_pair}, we show sample outputs of models using unpaired, paired and joint training approaches respectively. Due to the lack of supervision, the unpaired training can hardly recover the realistic textures. By adding paired training samples, the model is capable to generate more realistic images with well-preserved details.

\subsection{Loss Functions}
In this section, we describe the loss functions we used for the entire network. Pose Conditional GAN losses are used in both unpaired and paired training. Reconstruction loss, perceptual loss and style loss are used only for paired training.

\textbf{Pose Conditional GAN loss.}
We propose to use the Pose Conditional GAN to penalize whether the pose of the generated images are align with target person and whether the generated images are realistic. The input of the discriminator of the $PGAN$ is the concatenation of the person image $P$ and the dense pose $P_D$ for the person image. The $P'$ and $P'_D$ indicate the generated person image and the target dense pose for the generated image respectively. The loss function of the $PGAN$ is formulated as below:
\begin{equation}
\mathcal{L}_{PGAN} = \mathbb{E}[log\mathit{D(P, P_D)}] + \mathbb{E}[log\mathit{(1 - D(P', P'_D))}].
\end{equation}

Note that in the paired case, the ground truth of the transferred image is the same as the input $P$ because the model clothes is the same as the input person clothes. This allows us to use these paired images to train the model with the following losses.

\begin{figure*}[t]
    
    \includegraphics[width=\textwidth]{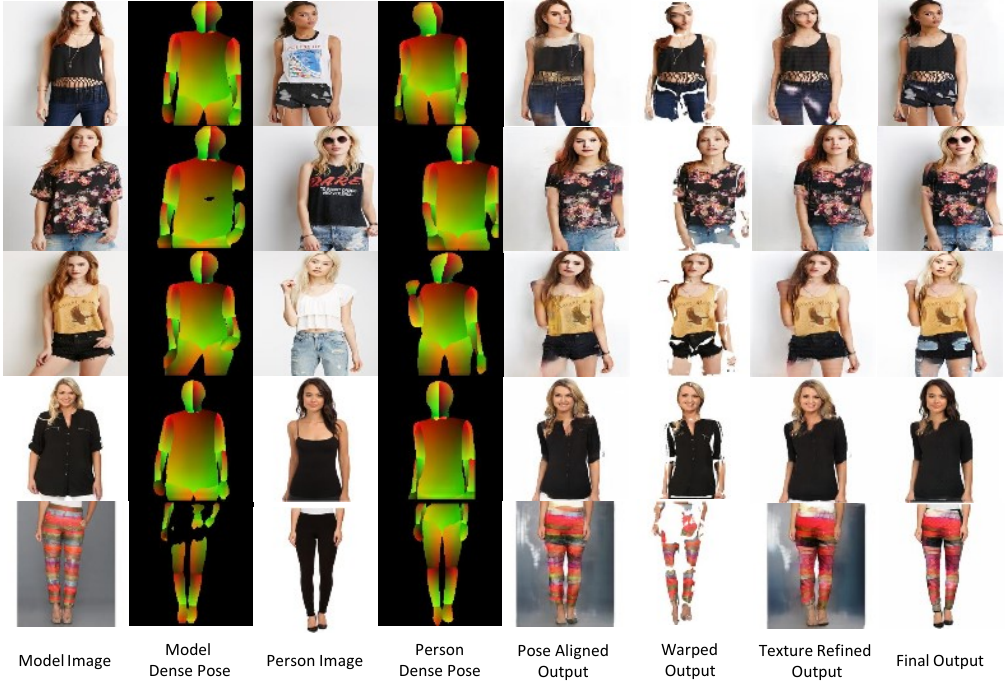}
    
    \caption{Intermediate results of M2E-TON on mini Deep Fashion, mini MVC Women Tops and mini MVC Women Pants datasets.
    First three rows are from mini Deep Fashion dataset, the fourth row is from mini MVC Women Tops dataset, and the fifth row is from mini MVC Women Pants dataset.}
    \label{abalation}
\end{figure*}

\begin{figure*}[t]
\centering
    \includegraphics[width=0.85\textwidth]{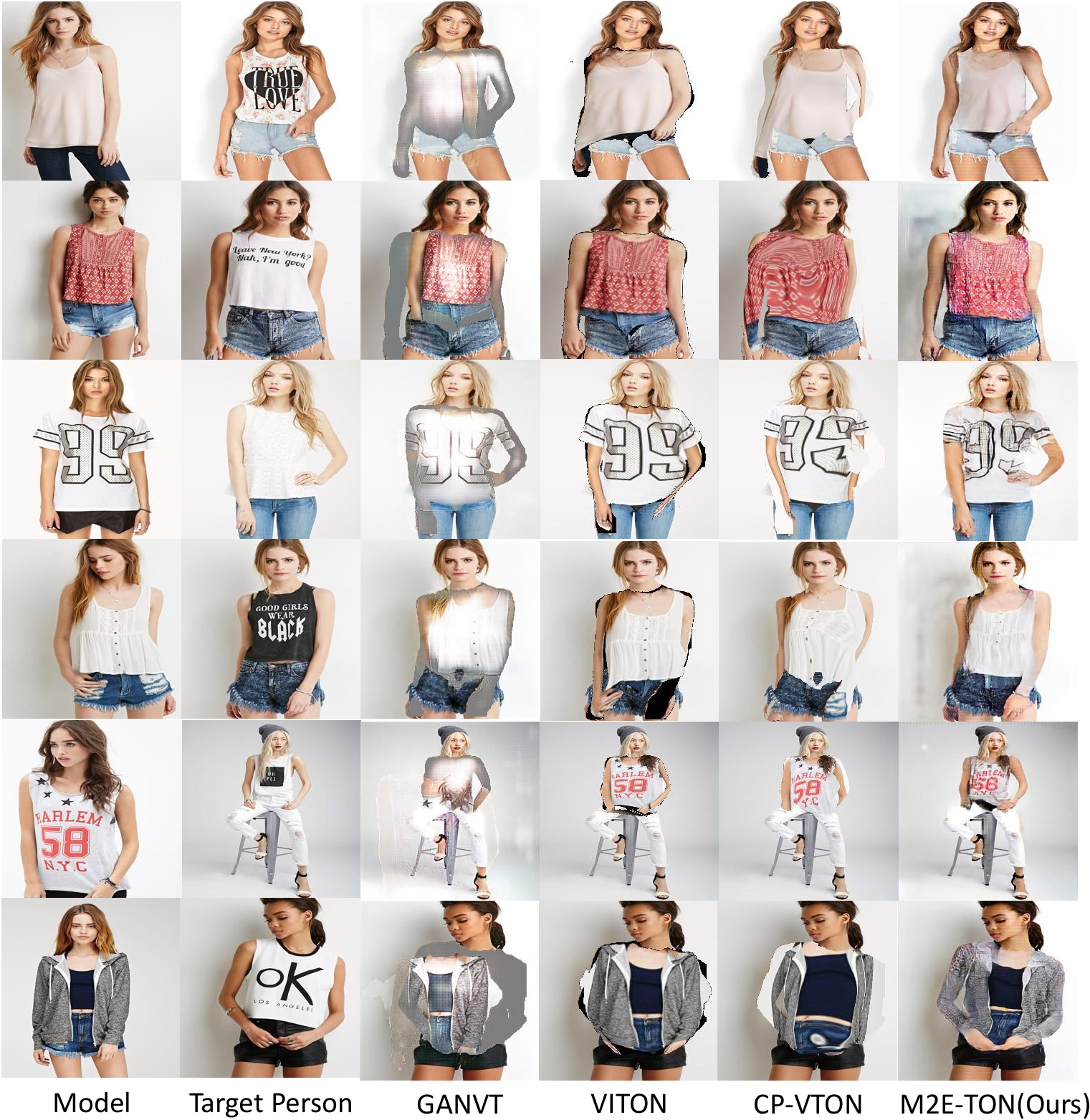}
    \caption{Qualitative comparisons of different methods. Our final results show the ability to fit the model's clothes to the target person by generating smooth images while still preserving the characteristics of both clothes and the target person.}
    \label{compare}
\end{figure*}

\textbf{Reconstruction Loss.} In paired training, we use the $\ell$1 and $\ell$2 distance between the generated image ($P'$) and the ground-truth image ($P$): $\| P' - P \| _{1}$, $\| P' - P \| _{2}$. These losses may result in blur images but are important for the structure preservation.

\textbf{Perceptual Loss.} Similar to other paired training methods, we use a perceptual loss to capture the high-level feature differences between the generated image and the corresponding real image. We extract the feature for both $P'$, $P$ by the VGG16 network pre-trained on ImageNet image classification task. We extract 7 different features $\Phi_i$ from relu1\_2, relu2\_2, relu3\_3, relu4\_3, relu5\_3, fc6 and fc7. Then we use the $\ell$2 distance to penalize the differences:

\begin{equation}
\mathcal{L}_P(P', P) = \sum_{i=1}^{n} \| \Phi_i(P') - \Phi_i(P) \| _{2} .
\label{energy}
\end{equation}

\textbf{Style Loss.} Inspired by \cite{gatys2015texture, gatys2015neural}, we penalize the style difference between generated image and ground-truth image in terms of colors, textures, etc. To achieve this, we firstly compute the Gram matrix for the network features $\Phi_i$ from the pre-trained VGG16 network : 

\begin{equation}
\mathcal{G}_i(x)_{c,c'} = \sum_{w}  \sum_{h}\Phi_i(x)_c \Phi_i(x)_{c'} ,
\label{energy}
\end{equation}
where $c$ and $c'$ are feature maps index of layer $i$, $h$ and $w$ are horizontal and vertical pixel coordinates. Then we calculate the  $\ell$2 distance for the Gram matrix gained on each layer $i$ between the generated images $P'$ and ground-truth images $P$:

\begin{equation}
\mathcal{L}_{Style}(P', P) = \sum_{i=1}^{n} \| \mathcal{G}_i(P') - \mathcal{G}_i(P) \| _{2} .
\label{energy}
\end{equation}

\section{Experiments}
    
\subsection{Datasets}
We experimented our framework on a subset of Deep Fashion dataset \cite{liu2016deepfashion}, mini MVC Women Top dataset \cite{liu2016mvc}, and MVC Women Pants dataset \cite{liu2016mvc}.
For Deep Fashion dataset, we select, 3256 image pairs for our unpaired training, 7764 image pairs for our paired training and 4064 image pairs for testing. For mini MVC Women Tops dataset, we select 2606 image pairs for our unpaired training, 4000 image pairs for our paired training and 1498 image pairs for testing. For mini MVC Women Pants dataset, we select 3715 image pairs for unpaired training, 5423 image pairs for our paired training and 650 image pairs for testing. All these images are resized to 256 $\times$ 256.

\subsection{Implementation Details}

We use Adam optimizer with $\beta_1$ = 0.5, $\beta_2$=0.99 and a fixed learning rate of 0.0002 for both unpaired training and paired training. We trained 80 epochs, 50 epochs and 50 epochs for the three sub-networks respectively.

\subsection{Baseline Methods}
We consider the following baseline methods to validate the effectiveness of our proposed M2E-TON. 

\textbf{GAN Virtual Try-On (GANVT).} We compare our method with GANVT \cite{kubo2018generative} which requires the product image as a condition in their GAN based network. In this case, we use the model clothes RoI to generate clean product image.

\textbf{VIirtual Try-On Network (VITON).} VITON \cite{han2017viton} is a virtual try on method which is able to generate synthesized images from the product clothes images and the clothing-agnostic person representation. We would like to highlight that the VITON is not a strict baseline method for our proposed M2E-TON due to the necessary usage of the clean product images which is not required for our method. Moreover, hardly obtained paired training data with both clean product images and the the person with the desired images is required for the VITON training which is also not required for us. For comparison, we conduct the experiments to compare with the TPS \cite{belongie2002shape} which is used in VITON to transform the product image shape to fit the human pose. We extract the model clothes RoI and apply TPS transformation to this extracted region, and merge the transformed clothes to the target person image.

\textbf{Characteristic-Preserving Virtual Try-On Network (CP- \\ VTON).} CP-VTON \cite{wang2018toward} is a state-of-the-art virtual try-on model which is proposed to try on the product clothes images to the desired person while preserving the detail clothes information. Similar to the VITON, the clean product images and the paired training data is used while it is not required for our M2E-TON. To compare with the CP-VTON, we conduct experiments in using the GMM warping on the model image conditioned on the dense pose estimation of both model and target person images while it is conditioned on the product images and the human pose estimation in the CP-VTON. Then we extract the transformed RoI region and merge it with the target person image. 

Note that all of the aforementioned methods require clean product images and paired trained data that are not available in our datasets. We use the RoI region of warped model images as the substitutes of the clean product images in their methods to reproduce the results. However, it is still hard to perform fair comparison.

\begin{figure}[]
\centering
    \includegraphics[width=0.4\textwidth]{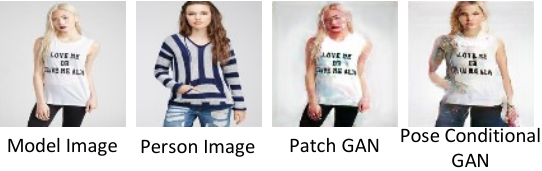}
    \caption{Comparison of patch GAN and our proposed pose conditional GAN in the unpaired training for the PAN. Our proposed pose conditional GAN is able to not only generate the real images but also align the generated images with target pose.}
    \label{compare_patch_GAN}
\end{figure}

\begin{figure}[]
\centering
    \includegraphics[width=0.4\textwidth]{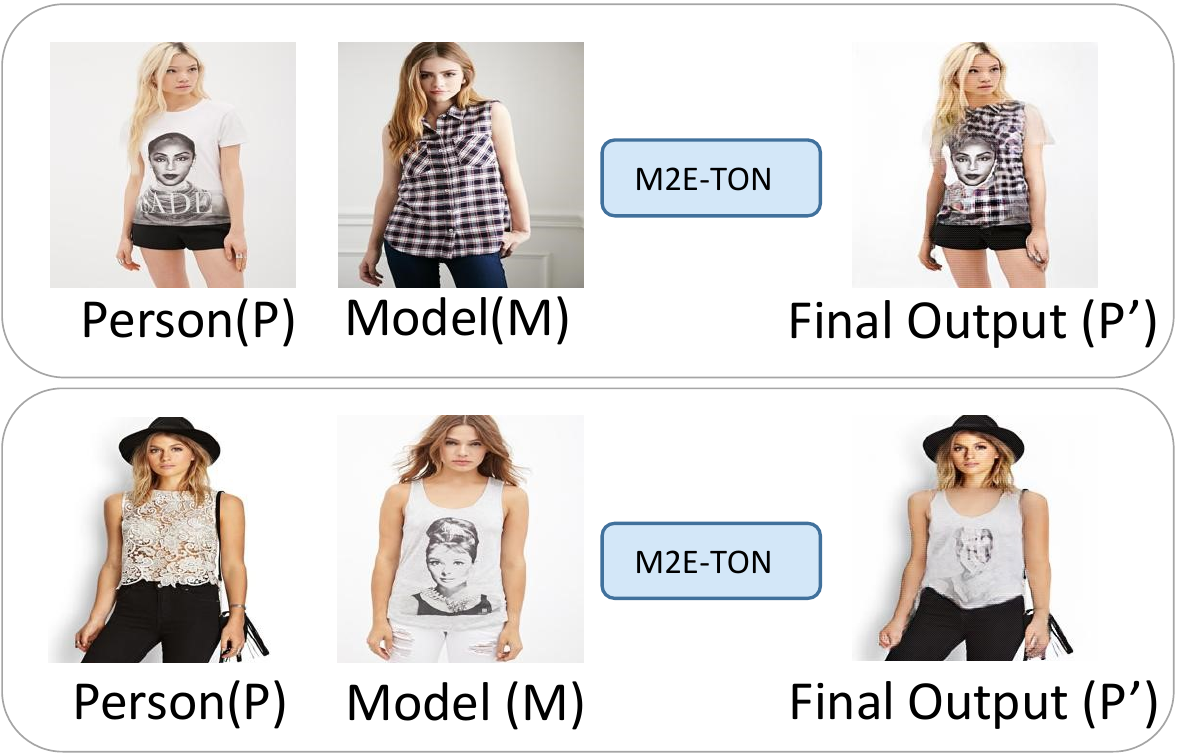}
    \caption{Failure cases of our method. Our method typically fails in the cases where there is a head pattern on the clothes.}
    \label{failure}
\end{figure}

\begin{table}[]
 
\caption{Quantitative Results for our M2E-TON. The User Study shows most of the users choose our final results as the best clothes visual try-on result. }
\centering
\label{table_result}
\begin{tabular}{l|c}
\hline
methods       & User Study \%
\\ \hline \hline
GAN-VT          &   0.8          \\ 
VITON     &  8.5          \\
CP-VTON       &      7.0       \\ 
Ours(Final)     &      83.7       \\ \hline
\end{tabular}
\end{table}
\subsection{Qualitative Results}
\textbf{Intermediate results.}
The generated results from different stages of M2E-TON model are shown in Figure \ref{abalation}.

\textbf{Comparison.} 
Figure.~\ref{compare} presents a visual comparison with the above-mentioned baseline methods. Our model shows great ability to fit the model's clothes to the target person by generating smooth images while still preserving the characteristics of both clothes and the target person. 

\textbf{Effectiveness of Pose Conditional GAN.}
We compare our proposed pose conditional GAN with the original patch GAN \cite{isola2017image} in Figure.~\ref{compare_patch_GAN}. If we use an unconditioned discriminator to distinguish real and fake patches in the unpaired training, the poses may not be well-aligned to the target poses since there are no pose-specific constraints. In comparison, our pose conditional discriminator can conditionally penalize the misalignment between the poses and the generated person images to guide pose transformation more effectively.

\subsection{Quantitative Results}

\textbf{User Study.} As shown in Table \ref{table_result}, we conduct a user study on mini Deep Fashion dataset to compare the quality of try-on images with a set of baselines. We select 20 users to choose the best image from the results generated by the four different approaches. For each user, we randomly select 100 images from the whole testing dataset. Note that human evaluation needs users to choose the image which looks the most realistic and preserves the most detail information. From the user study result, shown in Table \ref{table_result}, we can see that most people prefer to regard the virtual try-on images from our proposed M2E-TON as the most realistic try-on images. 

\subsection{Typical Failure cases}

Figure \ref{failure} shows some failure cases of our method. Our method typically fails in the cases where there is a head pattern on the clothes. The networks tend to treat this head pattern as head of the human instead of clothes region.

\section{Conclusion}
In conclusion, we have presented a novel virtual try-on network M2E-TON which can transfer desired model clothes to target person images automatically. We firstly utilize a pose alignment network (PAN) to align the model image with the target person image. A Texture Refinement Network (TRN) is proposed to enhance the characteristics in the aligned model image. With the refined apparel image, we fit the clothes to the target person with our Fitting Network (FTN). Moreover, we have proposed an Unpaired-Paired Joint Training strategy with the pose conditional GAN to ease the issue where the needs of high cost paired training data (the target person with desired model clothes). Our results demonstrate that the proposed method is able to generate photo-realistic visual try-on images.

~\\

\textbf{Acknowledgements.} This research was partially carried out at the Rapid-Rich Object Search (ROSE) Lab at the Nanyang Technological University, Singapore. The ROSE Lab is supported by the National Research Foundation, Singapore, and the Infocomm Media Development Authority, Singapore. This research is also supported by MoE Tier-1 Grant (RG28/18) of Singapore and NTU DSAIR Center. G. Lin's participation was supported by the National Research Foundation Singapore under its AI Singapore Programme [AISG-RP-2018-003] and the MOE Tier-1 research grant [RG126/17 (S)].

%
\bibliographystyle{ACM-Reference-Format}
\bibliography{egbib.bib}

\end{document}